\documentclass[a4paper, 10pt, conference]{ieeeconf}    

\usepackage[a4paper,
            left=054pt,
            right=37pt,
            top=104pt,
            bottom=54pt]{geometry} 

\usepackage{graphics} 
\usepackage{epsfig} 
\usepackage[citestyle=ieee]{biblatex} 
\usepackage{float}

\IEEEoverridecommandlockouts  

\begin{document}

\title{\vspace{18pt}\LARGE\textbf{{Multimodal Proximity and Visuotactile Sensing With a Selectively Transmissive Soft Membrane}}
\vspace{-3mm}
\thanks{J. Yin, G.M. Campbell, J. Pikul, and M. Yim are with the Department
of Mechanical Engineering and Applied Mechanics and the GRASP Laboratory at the University of Pennsylvania, Philadelphia, PA 19103 USA. \{jessyin, gcampbel, pikul, yim\} @seas.upenn.edu.}
 \thanks{This work was supported by the National Science Foundation Graduate Research Fellowship Program under Grant No. 2020295381 and NSF \#1935294.} 
}
\vspace{-10mm}
\author{{Jessica Yin, Gregory M. Campbell, James Pikul, and Mark Yim}}



\maketitle
\begin{abstract}
The most common sensing modalities found in a robot perception system are vision and touch, which together can provide global and highly localized data for manipulation. However, these sensing modalities often fail to adequately capture the behavior of target objects during the critical moments as they transition out of static, controlled contact with an end-effector to dynamic and uncontrolled motion. In this work, we present a novel multimodal visuotactile sensor that provides simultaneous visuotactile and proximity depth data. The sensor integrates an RGB camera and air pressure sensor to sense touch with an infrared time-of-flight (ToF) camera to sense proximity by leveraging a selectively transmissive soft membrane to enable the dual sensing modalities. We present the mechanical design, fabrication techniques, algorithm implementations, and evaluation of the sensor's tactile and proximity modalities. The sensor is demonstrated in three open-loop robotic tasks: approaching and contacting an object, catching, and throwing. The fusion of tactile and proximity data could be used to capture key information about a target object's transition behavior for sensor-based control in dynamic manipulation.
\end{abstract}

\section{Introduction}
Approaches to perception for robot manipulation have largely mimicked the human form, focusing on the development and integration of vision sensors far from the target object and compliant tactile sensors embedded in the end effector. However, robots still struggle to achieve dexterous and dynamic manipulation capabilities comparable to humans, particularly in the case of deformable objects. This can be largely attributed to uncertainties stemming from an imperfect perception of the target object \cite{billard2019trends}. The accuracy of the object's pose estimation can make the difference between success and failure, which can be seen in ``basic" tasks such as grasping, but is further amplified in  dexterous and dynamic tasks that lack simple contact models and quasi-static assumptions to inform the interaction \cite{ruggiero2018nonprehensile}. \par
While vision sensors provide rich data about the environment and can be used to localize the target object within it, the localization estimate is not very precise -- typically within a few centimeters around the object. Additionally, vision sensors are frequently occluded by robot arms as they reach towards the target or by clutter in the environment. An obvious short-term solution may be to simply add more cameras, but complete coverage of the target and workspace is not guaranteed even with multiple cameras and is not practical for real-world environments. \par
Tactile sensors on robot fingers and palms have been explored as a potential solution to provide more precise data about the object during contact, such as location and forces \cite{li2020review}. These sensors are usually designed to have mechanical compliance for increased robustness to unexpected contacts and greater functionality with the irregular or delicate geometries found in everyday objects. However, tactile sensors are only useful once the object is already in contact with the end effector, which may not be sufficient for tasks that require bringing the object in and out of contact, such as dynamic reorientation. \par
This points to a fundamental gap in a perception pipeline that only uses vision and touch. Closing this perception gap is necessary to create a robust perception pipeline that will allow robots to tackle more difficult manipulation tasks. A potential solution to address this gap is adding a proximity sensing modality, which can be defined as sensing within a short distance range originating from the locations of the tactile sensors \cite{navarro2021proximity}. Proximity sensing can provide the precise localization data that vision sensors lack and information about pre- and post-contact behavior that is difficult to predict due to complex frictional dynamics. \par
 \begin{figure}[t!]
  \centering
  \includegraphics[width=\linewidth]{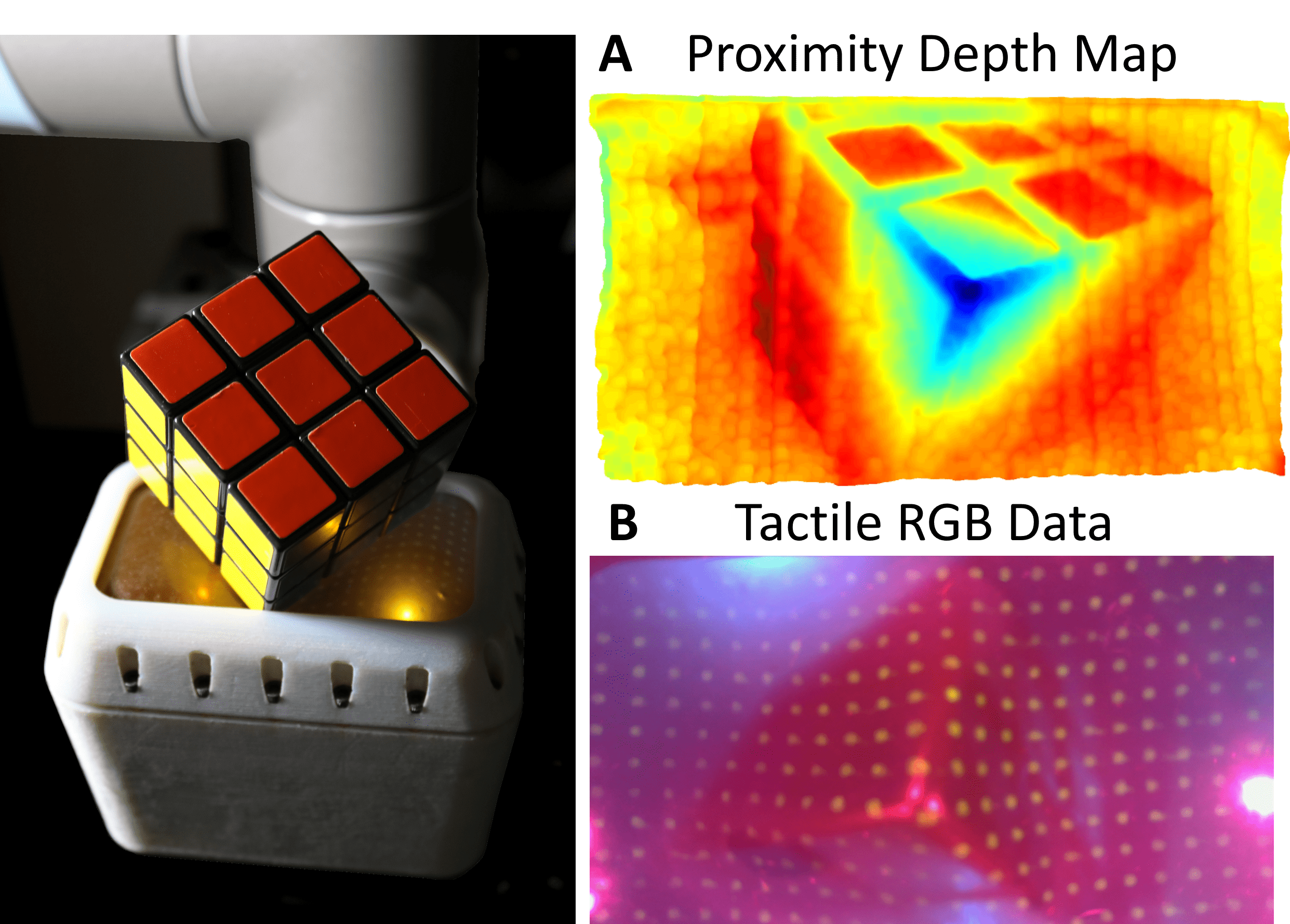}
  \caption{A. Proximity depth map from internal depth camera. B. Image from internal RGB camera for tactile data. }
  \vspace{-3mm}
  \label{fig1}
  \end{figure}
  
  \begin{figure*}[t!]
    \centering
    \includegraphics[width = \textwidth]{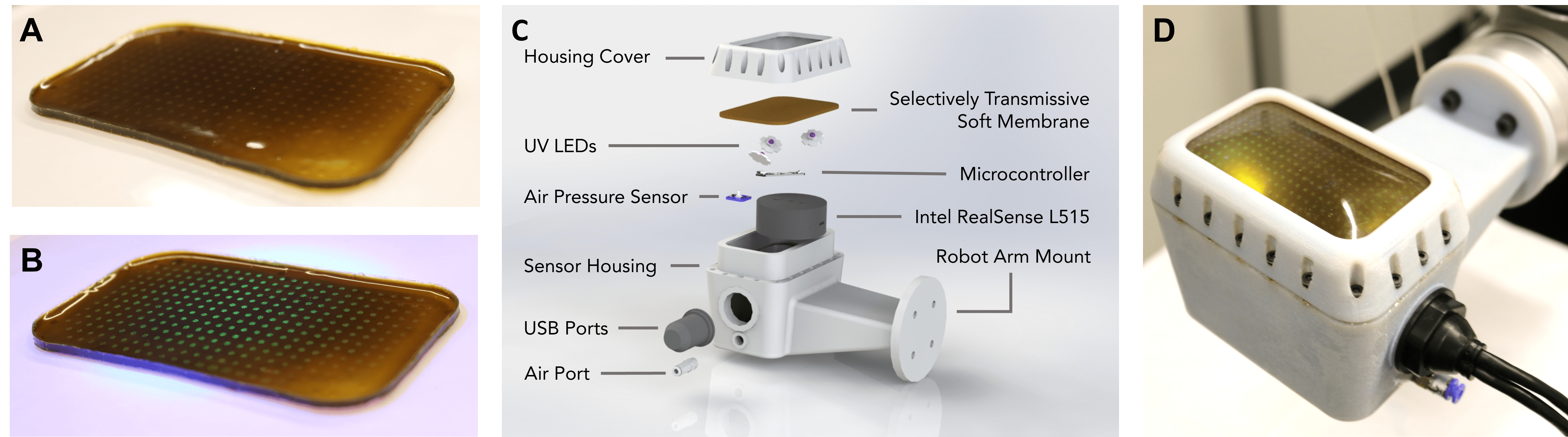}
    \caption{A. Soft membrane in ambient room light. B. Soft membrane with UV-phosphorescent dot grid pattern activated by 365nm UV light. C. Overview of sensor system. D. Sensor mounted on UR10 robot arm.}
    \vspace{-3mm}
    \label{fig:fig2}
\end{figure*}
In this paper, we propose a novel multimodal proximity and visuotactile sensor that provides simultaneous tactile and proximity depth data. The sensor is able to detect contact over an inflated 96mm by 54mm elastomer membrane with an RGB camera (960x540) and air pressure sensor, while providing depth data with an infrared (IR) ToF camera (640x480) at a synchronous sampling rate of 30Hz. An infrared-transmissive and visibly translucent elastomer membrane, embedded with UV-phosphorescent particles, enables the simultaneous reading of visible tactile data on the membrane and IR proximity data. We introduce a sensor fusion algorithm that uses both the RGB image and depth image to correct for the effect of the embedded particles in the depth image, and evaluate the depth data up to 100mm from the sensing surface. The sensor is integrated into an end-effector and mounted on a UR10 (Universal Robots) robot arm and demonstrated with the following open-loop tasks: approach and contact, catching, and throwing. 

\section{Related Work}
Visuotactile sensors are a common tactile sensing strategy due to their high resolution, large coverage, and relatively easy fabrication process. These sensors contain a camera that observes a visual pattern on the internal surface of a soft membrane \cite{shimonomura2019tactile}. The soft membrane deforms when in contact with an object and the visual pattern distorts. This pattern distortion is captured by the camera and can be used to estimate tactile data, such as shear displacement, forces, and contact area. Most visuotactile sensors use a dot pattern on the membrane to track deformation and frustrated total internal reflection to estimate depth for information about a target object \cite{yuan2017gelsight, donlon2018gelslim, lambeta2020digit, padmanabha2020omnitact, ward2018tactip}. Another approach is to use a ToF depth camera to directly sense deformations of the membrane's surface \cite{alspach2019soft}. However, these sensing mechanisms fundamentally prohibit the integration of a proximity modality, and these sensors can struggle to resolve ambiguous tactile imprints to determine an object's pose.  \par
There are a wide range of proximity sensing strategies, from optical fibers \cite{konstantinova_force_2015} to single time-of-flight sensor point measurements \cite{hellebrekers2018liquid}. Proximity sensing has mostly been implemented on the fingertips of robotic grippers for pre-grasp object detection \cite{yin2020closing} and improved grasping \cite{sun_improving_2018}. However, these proximity sensors are limited in spatial resolution, which prevent tasks such as object recognition and tracking.\par
FingerVision uses stereo vision to observe a clear elastomer surface marked with black dots \cite{yamaguchi2016combining}. This design, however, compromises between tactile and proximity resolutions; adding more tactile dots requires occluding the proximity vision. See-Through-Your-Skin achieves modulated transparency via the two-way mirror effect for visual proximity and tactile data \cite{hogan2021seeing}. However, it cannot provide both simultaneously and has a limited deformation range due to its rigid platform. Multimodal proximity and tactile sensing has also been achieved through magnetic \cite{hellebrekers2020localization} and capacitive sensing \cite{rocha2017soft}, although the proximity sensing depends on the target object's capacitive properties or placement of magnetic stickers around the workspace and object. \par
We build upon previous work in multimodal proximity and visuotactile sensing and extend it with high spatial resolution depth data and synchronized, real-time tactile and proximity data. Furthermore, our tactile and proximity modalities cause minimal to no interference with the other, leading to uncompromised spatial resolution of each modality.
\section{Design and Fabrication}
\subsection{Selectively Transmissive Soft Membrane}
We designed the soft membrane for selective transmission to achieve the following: (1) allow the infrared light (860nm) emitted by the time-of-flight camera to pass through, (2) block most of the visible light (400nm-700nm) from the external environment, and (3) enable the activation of the phosphorescent, light-emitting particles ($\sim$500nm) on the inner surface from internal UV LEDs (365nm). Blocking external visible light enhances the visual contrast with the green-colored phosphorescent particles (Figure \ref{fig:fig2}A, \ref{fig:fig2}B), facilitating the facile application of off-the-shelf OpenCV algorithms for tracking \cite{bradski2000opencv}. Additionally, the membrane is designed to be physically resilient for repeated use in contact-rich interactions, while providing a highly compliant contact surface. The thickness of the membrane can be decreased for greater infrared light transmissivity, but at the cost of reduced physical robustness and opacity to external light. \par
We fabricate the membrane in layers; each silicone elastomer layer fully cures prior to pouring the next layer. The base silicone elastomer (Ecoflex 00-30; Smooth-On) has an attenuation of $\sim$10db/cm at 860nm \cite{zhao2016optoelectronically}. With a final thickness of 1mm, the membrane therefore transmits 91\% of the emitted infrared light from the depth camera. The first layer consists of Ecoflex 00-30 mixed by hand with a dye solution (Epolight 7276B; Epolin, dissolved in chloroform, 5.325g/L concentration) in a 15:1 (mL) elastomer to dye solution ratio. The dye solution is visibly opaque and infrared transmissive. We pour 8.25g of the dyed elastomer into an Ease-Release-coated laser cut mold and place it in a vacuum degassing chamber for 10 minutes. Then, the mold is placed on a hot plate and heat-cured at 100\textdegree C for 10 minutes. \par
The next layer consists of the UV-phosphorescent particles in a dot grid pattern. The UV-phosphorescent particles are made of Cu:ZnS (copper doped zinc sulfide, 35 microns; Technoglow). We mixed 0.2g of Cu:ZnS with 2g of Ecoflex 00-30 by hand. We laser cut a 0.508mm stencil made of clear PVC to the shape of the membrane and desired dot grid pattern (1mm diameter, 4mm uniform spacing, 328 total dots). The first layer of the membrane is then removed from the mold and placed onto a glass plate. We press the stencil onto the membrane to remove air bubbles and the Cu:ZnS elastomer mixture is spread onto the surface with a q-tip. The stencil is removed after 15 seconds and the glass plate with the membrane cures on a hot plate at 100\textdegree C for 10 minutes.\par
The final layer evens out the protrusions from the dot grid layer and leaves a slightly matte finish to reduce specular reflections from the infrared and UV lights. We mix Ecoflex 00-30 with NOVOCS Matte silicone solvent (Smooth-On) in a 3:6 (g) solvent to elastomer ratio and degas for 10 minutes. Finally, we pour 4g of the mixture onto the membrane and cure at 100\textdegree C on a hot plate for 10 minutes. 
   \begin{figure}[t!]
  \centering
  \includegraphics[height=3in]{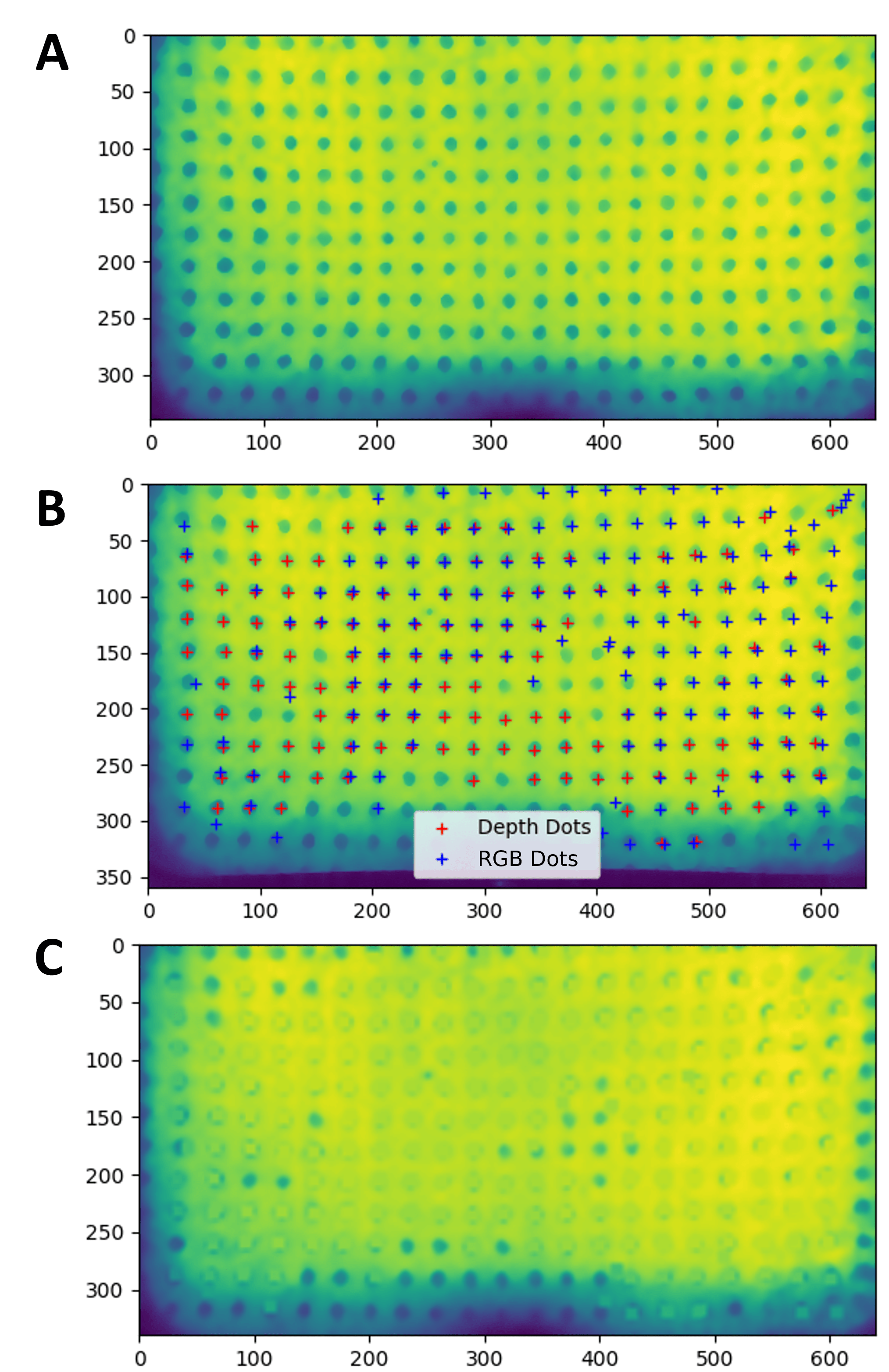}
  \caption{A. Depth map before dot correction algorithm is applied. B. Dots detected by blob detector in the depth and RGB images. C. Depth map after dot correction is applied.}
  \label{Dot_Correction}
  \vspace{-5mm}
  \end{figure}
 
\subsection{Internal Electronics}
We chose to use the Intel Realsense L515 because it provides integrated RGB and ToF depth cameras, as well as the ability to adjust the ToF laser power to bring the minimum sensing range to approximately 50mm for short range sensing. The field of view (FOV) of the RGB camera is 70\textdegree {} by 43\textdegree and the FOV of the depth camera is 70\textdegree {} by 55\textdegree. Because the FOVs don't exactly align, the active sensing area in this work only consists of the overlapping regions of both FOVs. The cameras output data through the same USB-C port, which is connected to an airtight USB-A 3.0 port that goes through the sensor housing. The internal air pressure sensor samples data at 30Hz and is connected to a microcontroller. 
We inflate the membrane to a gauge pressure of 0.02PSI to reduce the specular reflection of the internal UV and IR lights. \par
We soldered three UV LEDs (365nm) onto aluminum heat sinks and connected them in series with 50m$\Omega$ resistance. The LED circuit is connected to the direct USB power output pin on the microcontroller, which provides 2.1A. The microcontroller is connected to the USB-A port that goes through the sensor housing. An overview of the internal components of the sensor system is shown in Figure \ref{fig:fig2}C.

\subsection{Sensor Housing}
The 3D printed (DraftGray, Objet 30 Prime; Stratasys) sensor housing consists of the main sensor chamber and the cover. Threaded inserts are glued with epoxy resin (Gorilla 2 part epoxy; Gorilla Glue) around the housing and the cover is attached with M3 screws. The lid and the screws clamp the membrane down and create an airtight seal. The main sensor housing also has ports for the push-to-connect tube fitting and the dual USB-A 3.0 port. The sensor housing includes a mounting plate for the UR10 arm (Figure \ref{fig:fig2}D).

\begin{figure*}[th!]
    \centering
    \includegraphics[width = \textwidth]{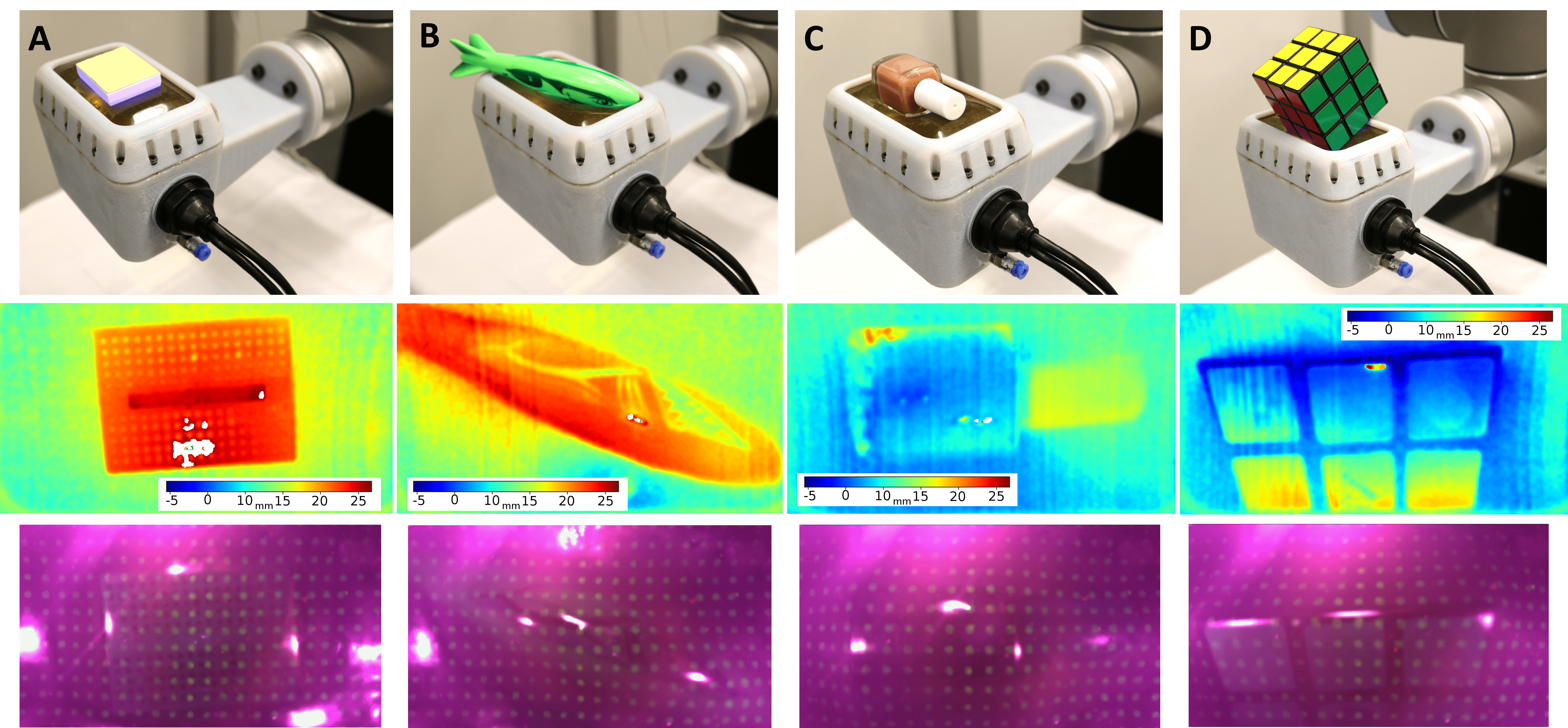}
    \caption{Top row: test object, middle row: corresponding proximity depth map, bottom row: corresponding tactile RGB image with phosphorescent dots for membrane tracking. A. Solderless breadboard. B. Shark torpedo. C. Nail polish. D. Rubik's cube.}
    \vspace{-3mm}
    \label{fig:testobjs}
\end{figure*}
\section{Sensor Characterization}
\subsection{Dot Correction Through Sensor Fusion}

While the membrane is designed to maximize transmission of the emitted infrared light for depth sensing, the UV phosphorescent dots do introduce some light scattering compared to an unpatterned region of the membrane. The dots are imperceptible in the depth map when an object is within approximately 40mm of the sensing surface and when it is in contact. This is potentially because the object is reflecting enough infrared light back such that scattering effects become negligible. However, a distinct dot pattern appears in the depth map when sensing objects far away from the sensor (greater than 40mm), with the dot-patterned regions appearing 2mm-4mm closer to the camera. We correct for the dot pattern in the depth map by fusing the depth and RGB data (Figure \ref{Dot_Correction}). Because the dots are always visible and actively tracked with the RGB camera, the RGB images can be used to apply corrections when the dots are affecting the depth images. \par
To map the dots in the RGB data to their correct location within the depth map, we first align the RGB frame to the depth frame. The RGB and ToF cameras are located less than 2cm apart and approximately within the same plane. We estimated the relevant transform matrix from calibration data to align the RGB and depth images. The transform matrix was then manually tuned based on the overlap of the transformed RGB image and depth data from a flat plane resting 100mm from the sensors. This tuned transform was found to be appropriate for all image alignments with object distances between 40mm and 100mm from the sensor.\par
After the visual image is aligned to the depth frame, we used the simple blob detector from OpenCV to identify and locate the dots in both the depth image and the aligned RGB image. Dots are assumed to exist in the superset of these two sets of detected dots. At each dot location, localized smoothing is applied based on the distance values of the neighboring points. We take the distance value from eight pixels (one from each cardinal and ordinal direction) and apply the average of these eight distances to a grid centered at the center of the dot. Finally, a global smoothing is applied with a Gaussian blur.

\subsection{Proximity Depth Sensing}

\begin{figure}[t!]
    \centering
    \includegraphics[width=0.8\linewidth]{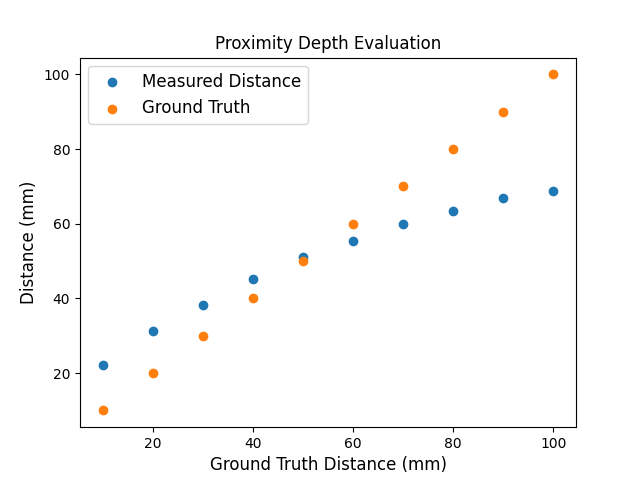}
    \caption{Measured distance compared to ground truth distance for a flat plane over a range of 10-100mm.}
    \vspace{-5mm}
    \label{fig:depth_error}
\end{figure}

In this section, we characterize the proximity depth sensing. The relevant settings for the Intel Realsense L515 are the following: laser power - 10, receiver gain - 18, digital gain - 1, minimum distance - 0mm, and all filters (confidence, decimation, noise) and pre-/post-processing sharpening turned off. These settings are kept consistent throughout this work. A test stand with discrete slots from 10mm-100mm in 10mm increments mounts a flat plane parallel to the sensing surface. Four sheets of white printer paper (92 brightness) cover the flat plane and encompass the entire FOV of the depth sensing. We apply the dot correction algorithm and average 20 consecutive frames for evaluation of the data. The average depth pixel value of the depth data has an $R^2$  = 0.725 fit with the ground truth distance (Figure \ref{fig:depth_error}). The lowest average error, 1mm, occurs at a distance of 50mm, while the largest average error, 31mm, occurs at a distance of 100mm. The sensor tends to both underestimate further distances and overestimate closer distances because the returned signals are generally weaker than its out-of-the-box calibration. The depth data shows spatially varying accuracy at further distances, particularly above the 50mm distance range, due to the convex nature of the laser power output \cite{intelrealsense}.\par 
Figure \ref{fig:testobjs} shows depth maps of objects placed on the surface of the sensor. We compared the depth maps to the significant dimensions of each object and found an average overall error of 4.3\%. The sensor showed poorer performance on curved surfaces, with the average error of 5.5\% and much better on edges, with an average error of 2\%. \par
Perhaps due to the on-chip confidence algorithm, an object in contact with the sensor causes the entire surface of the membrane to be sensed. This feature of the sensor should be tested with a more extensive range of objects, but it has remained consistent with the object dataset tested thus far. The reflectance properties of the object's surface has a significant effect on the quality of the depth map. For example, the black lines of the Rubik's cube overemphazies the separation of each square because it absorbs more of the infrared light, which is interpreted as further away from the sensor. On the other hand, the white center divider strongly reflects infrared light and thus shows up as much closer to the sensor than the rest of the breadboard. Additionally, object geometries such as edges can lead to deformations in the membrane such that the UV and/or infrared light is specularly reflected and cause outliers in the data, such as on the lower half of the breadboard in Figure \ref{fig:testobjs}A. 

\subsection{Tactile Sensing}
Tactile sensing is achieved by measuring the change in the internal air pressure and by tracking the motion of the dots on the internal membrane surface. The dots are detected in the RGB image with the simple blob detector and tracked with the Lucas-Kanade optical flow algorithms from OpenCV. The simple blob detector finds the center coordinates of each dot in each frame, and then the optical flow calculates the distance between its initial position and current position. To detect contact, the total flow velocity summed from all the dots act as a proxy for the magnitude of membrane deformation, and therefore total contact force. An RGB image of an uncontacted and inflated membrane initializes the optical flow and subsequent frames are compared to the uncontacted state. The air pressure sensor uses gauge pressure for contact detection.  \par 
Measuring both the internal air pressure and flow velocity for binary contact detection extends the range of contact that can be detected. The internal air pressure is more sensitive to contact and can detect forces below 100g, which are not sufficient to create an appreciable change in flow velocity. The flow velocity is particularly useful for detecting tangential forces and lateral motions of the object along the sensing surface, which may not produce significant changes in the air pressure. The sensitivity of the flow velocity contact detection can be tuned to detect different ranges of forces by changing the window size of the optical flow algorithm. \par

\section{Demonstrations}
We mounted the sensor to a UR10 robot arm and demonstrated tasks where it could be beneficial to use both proximity and tactile sensing modalities. Each task can be separated into ``pre-contact", ``during contact", and ``post-contact" stages that excite both sensing modalities. Furthermore, accurate perception of object behavior in all of these stages is critical for successful manipulation. Although the tasks are open-loop, they provide a first step towards sensor-based control. 
\begin{figure}[th!]
    \centering
    \includegraphics[width = \linewidth]{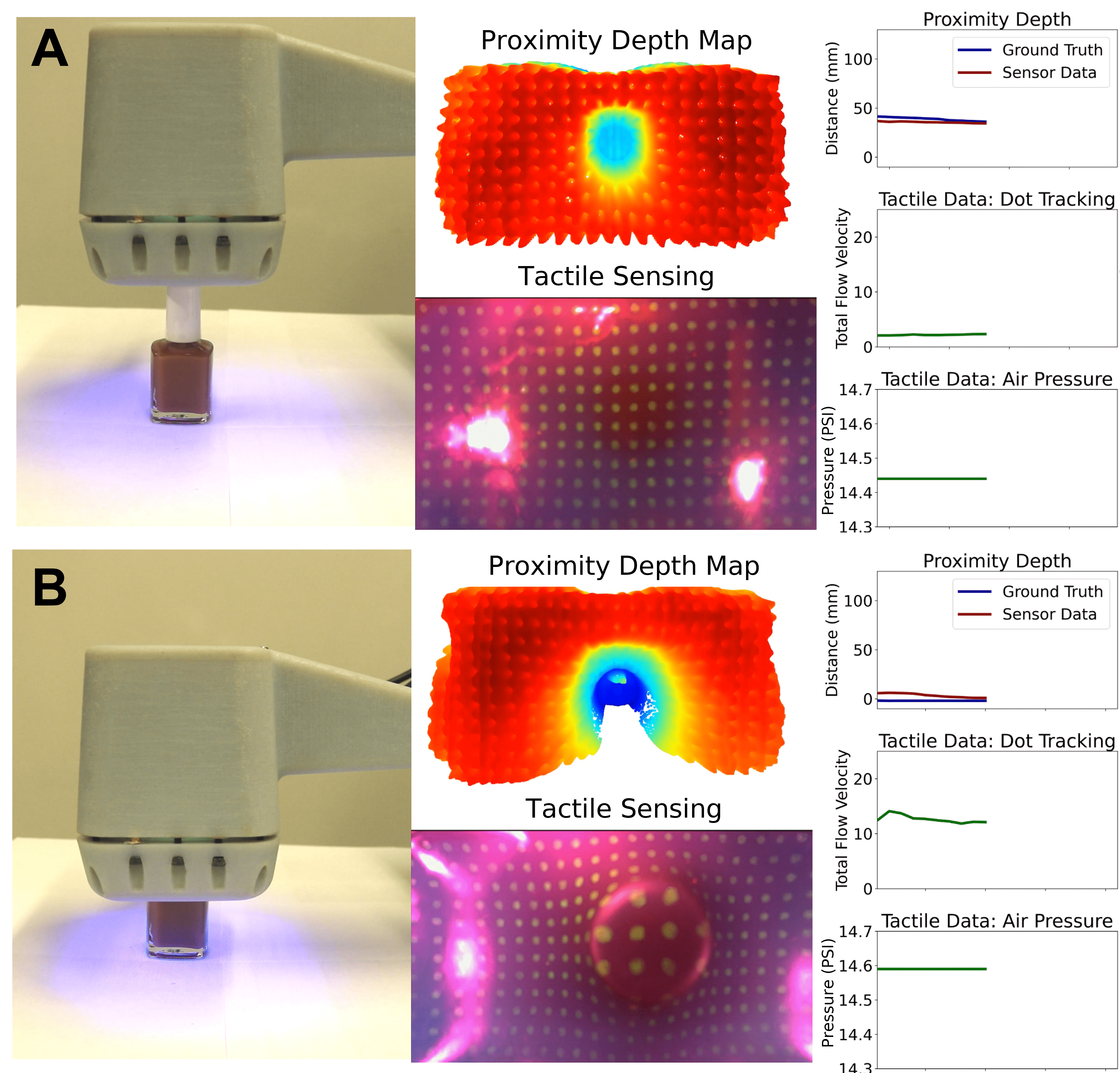}
    \caption{Proximity depth and tactile sensing data: A. Before the sensor makes contact with the nail polish. B. After the sensor makes contact with the nail polish. }
    \vspace{-3mm}
    \label{fig:contact}
\end{figure}
\begin{figure} [th!]
    \centering
    \includegraphics[width = \linewidth]{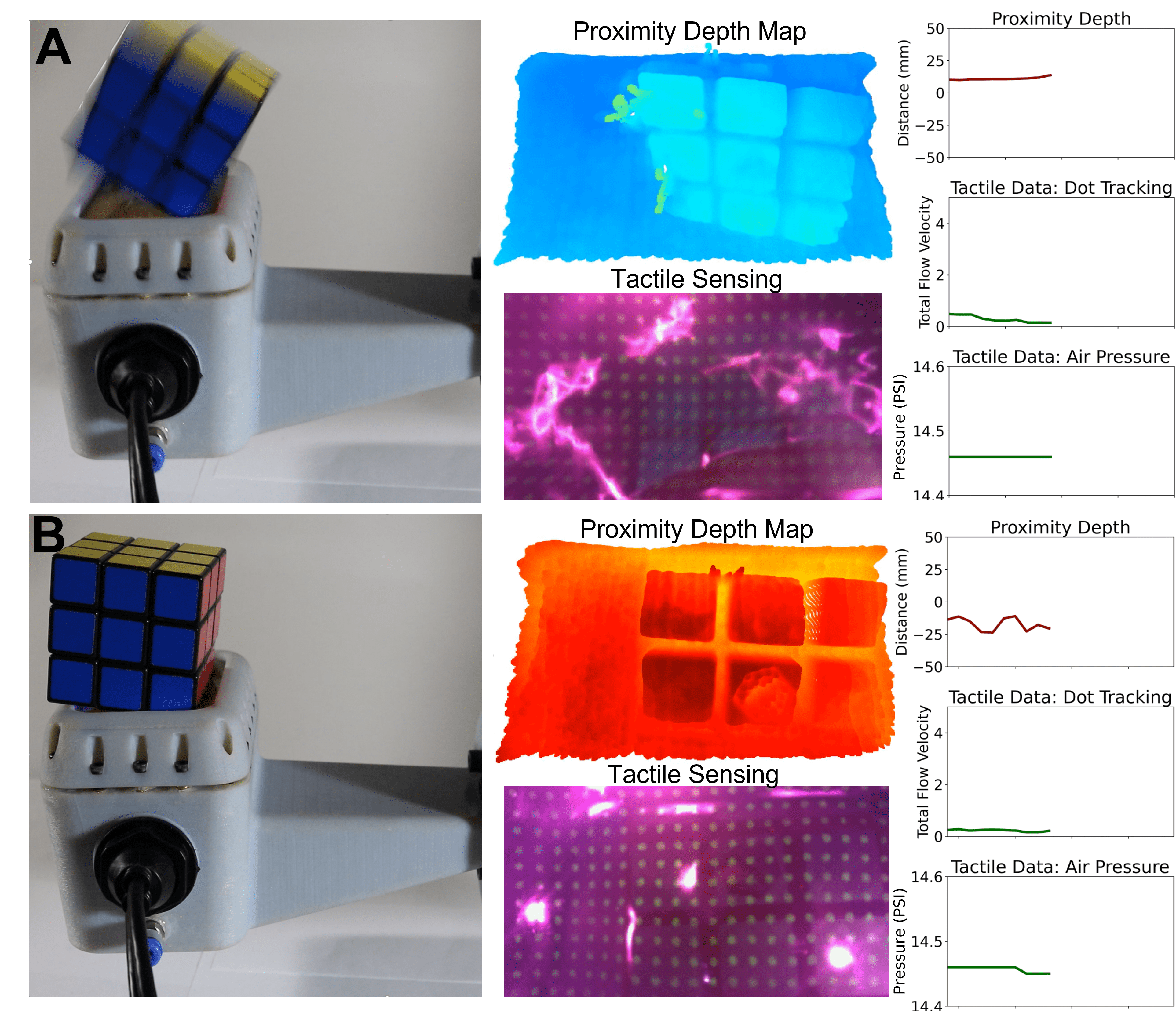}
    \caption{Proximity depth and tactile sensing data: A. Before the Rubik's cube makes contact with the sensor. B. After the Rubik's cube makes contact. }
    \vspace{-3mm}
    \label{fig:catch}
\end{figure}
\subsection{Approach and Contact}
The robot performs an approach and contact task with a bottle of nail polish placed on a table. The pose of the end effector from the robot arm provides the ground truth for the distance accuracy of the sensor. Data from the same approach and contact trial are shown in Figure \ref{fig:contact}. \par
The sensor's initial pose is oriented towards the nail polish, 40mm above the object (Figure \ref{fig:contact}A). The proximity depth map shows the top of the nail polish clearly in the center, while the tactile RGB image shows zero deformation of the sensing surface. The proximity plot on the left estimates the nail polish is 40mm away from the sensing surface, which is in excellent agreement with the ground truth. The tactile data plots on the right also show data that corroborates with the no-contact state: the dot tracking flow velocity is close to zero, and the air pressure has not changed from the initial air pressure. \par
The robot arm then moves the sensor towards the bottle at 50mm/s until it makes contact and protrudes 10mm into the sensing surface (Figure \ref{fig:contact}B). The proximity depth map shows the top of the bottle protruding into the surface and outlines the body of the bottle, while the tactile RGB image shows the circular top of the bottle producing a deformation of the sensing surface. The plots on the right show that the proximity estimation is in good agreement with the ground truth and that the nail polish is in contact with the sensing surface. The tactile data plots show a significant increase in dot tracking flow velocity from the membrane deformation, as well as a 0.15PSI increase in internal air pressure.
\par
The proximity and tactile sensing data also shows close agreement of when contact occurred: all plots on the right have synchronized time on the x-axes. The proximity depth modality shows good agreement with ground truth when the object is within 40mm of the sensor, but performs poorly beyond 40mm. 

\subsection{Catching}
To demonstrate catching, the arm-mounted sensor faces the ceiling and a Rubik's cube is dropped onto the sensor from a height of 80mm. Figure \ref{fig:catch} consists of data from the same catching trial, prior to and after contact. All x-axes on the rightmost plots are time-synchronized. \par
Figure \ref{fig:catch}A shows the Rubik's cube just before making contact with the sensor. The proximity depth map shows the angular orientation and the square features of the Rubik's cube, while the RGB tactile image shows no deformation. The plots on the right show that the Rubik's cube is 7mm above the sensing surface, and the air pressure and dot tracking flow velocity show little to no change from initial no-contact conditions. \par
Figure \ref{fig:catch}B displays data after the Rubik's cube settled onto the sensing surface. The proximity depth map and plot shows about half of the Rubik's cube protruding 22mm into the membrane. The tactile RGB image shows some deformation of the membrane surface, with a small increase in dot tracking flow velocity. The internal air pressure of the sensor decreased, potentially due to some air leaking after the impact of the Rubik's cube. 
\par
The proximity depth data senses the Rubik's cube 13 frames before it makes contact with the membrane. Both the proximity depth data and tactile sensing data match well qualitatively with a video recording of the experiment.
\begin{figure}
    \centering
    \includegraphics[width = \linewidth]{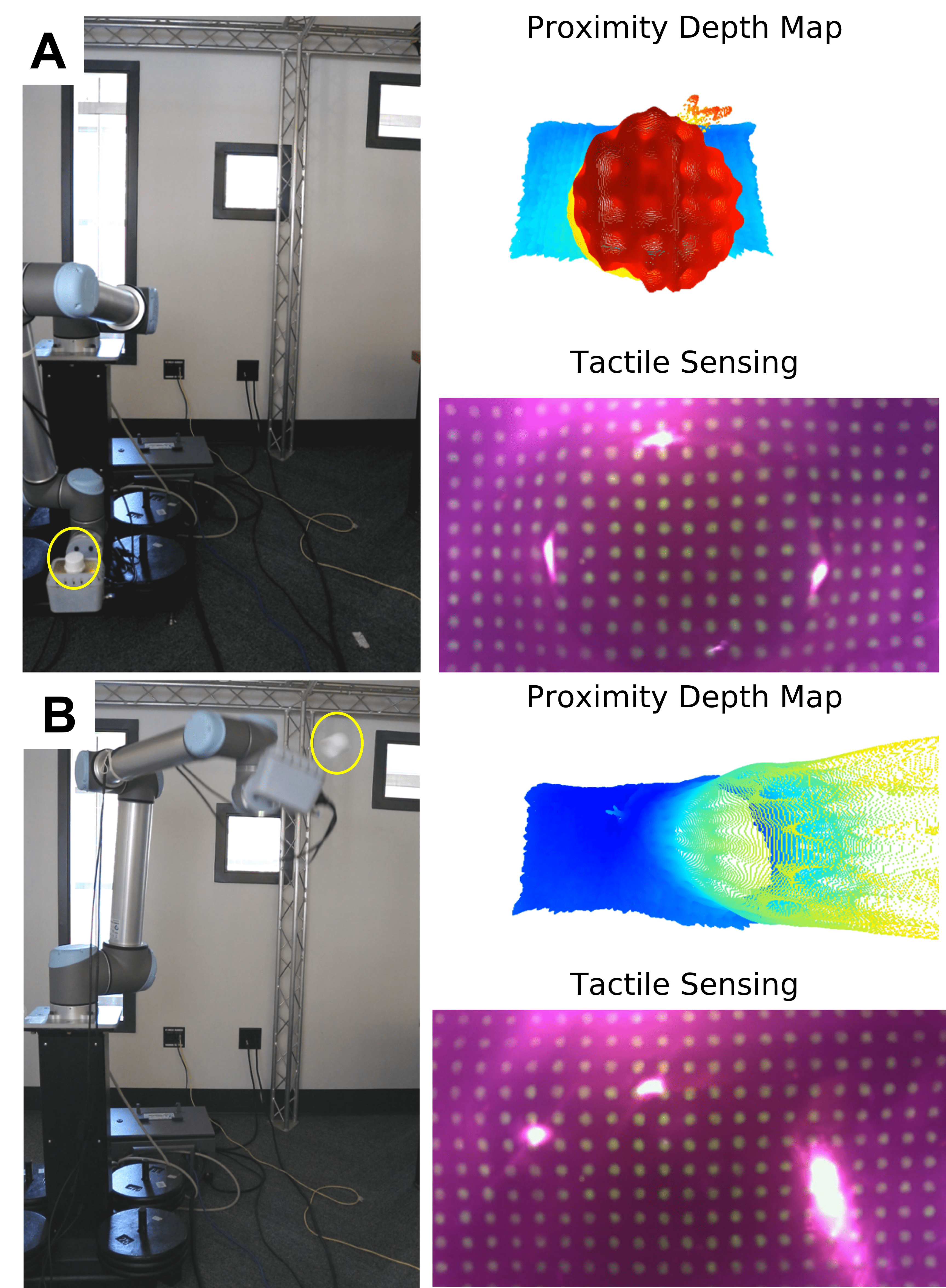}
    \caption{Proximity depth and tactile sensing data, with the target object circled in yellow: A. Prior to throwing the hex head cap. B. After throwing the hex head cap.}
    \vspace{-3mm}
    \label{fig:throw}
\end{figure}
\subsection{Throwing}
In this demonstration, the UR10 arm throws a 46mm diameter PVC hex head cap off the surface of the sensor (Figure \ref{fig:throw}). The final speed of the end effector is approximately 1.5m/s. Until the hex head cap is thrown, it remains in contact with the sensor, although we observe some lateral rocking during the wind-up trajectory in both the tactile and proximity data. Capturing this type of object behavior could be very useful in predicting object trajectories after a throw. The tactile and proximity data both show loss of contact at the end of the throw. After the hex head cap is thrown, the sensor captures 6 frames of the hex head cap's initial projectile motion. 

\section{Conclusions and Future Work}

In this study, we introduced a novel multimodal proximity depth and visuotactile sensor enabled by a selectively transmissive elastomer membrane. We presented the design and fabrication techniques for each component of the sensor, and we evaluated the proximity depth data across a distance range of 10mm-100mm. Both the binary contact detection and proximity depth modalities were tested with an object dataset consisting of nail polish, Rubik's cube, breadboard, shark torpedo, and PVC hex head cap. \par
We integrated the sensor into an end-effector to mount on a UR10 robot arm and demonstrated it in three open-loop tasks where the mixed modality of the sensor could provide an advantage. The demonstrations and quality of the data show potential for the application of this sensor to capture target object behavior before, during, and after contact in dynamic and dexterous manipulation tasks.
 \par
Moving forward, we plan to expand the development of tactile sensing functionality with monocular depth estimation of the membrane and the fusion of tactile and proximity data for contact patch and force estimation. Due to the nonlinear mechanics introduced by the use of a highly deformable elastomeric membrane, it is challenging to precisely relate force, deformation, and geometry. With force estimations and expanded tactile sensing capabilities, the sensor could be used to develop dynamic and dexterous control policies using proximity and tactile data. The current sensor package as an end-effector is suitable for future work in exploring the use of contact patches in closed-loop control. Because the sensor has a relatively large sensing area, it can also be used as a sensorized palm to investigate applications in in-hand dexterous manipulation.  

\section*{Acknowledgment}
The authors would like to thank Dr. Dinesh Jayaraman of the GRASP Laboratory, University of Pennsylvania for graciously allowing extensive use of the UR10 robot arm. \\




\section{References}
\printbibliography[heading=none]

@inproceedings{alspach2019soft,
  title={Soft-bubble: A highly compliant dense geometry tactile sensor for robot manipulation},
  author={Alspach, Alex and Hashimoto, Kunimatsu and Kuppuswamy, Naveen and Tedrake, Russ},
  booktitle={2019 2nd IEEE International Conference on Soft Robotics (RoboSoft)},
  pages={597--604},
  year={2019},
  organization={IEEE}
}

@article{li2020review,
  title={A review of tactile information: Perception and action through touch},
  author={Li, Qiang and Kroemer, Oliver and Su, Zhe and Veiga, Filipe Fernandes and Kaboli, Mohsen and Ritter, Helge Joachim},
  journal={IEEE Transactions on Robotics},
  volume={36},
  number={6},
  pages={1619--1634},
  year={2020},
  publisher={IEEE}
}

@article{billard2019trends,
  title={Trends and challenges in robot manipulation},
  author={Billard, Aude and Kragic, Danica},
  journal={Science},
  volume={364},
  number={6446},
  year={2019},
  publisher={American Association for the Advancement of Science}
}

@inproceedings{yamaguchi2016combining,
  title={Combining finger vision and optical tactile sensing: Reducing and handling errors while cutting vegetables},
  author={Yamaguchi, Akihiko and Atkeson, Christopher G},
  booktitle={2016 IEEE-RAS 16th International Conference on Humanoid Robots (Humanoids)},
  pages={1045--1051},
  year={2016},
  organization={IEEE}
}

@article{ward2018tactip,
  title={The tactip family: Soft optical tactile sensors with 3d-printed biomimetic morphologies},
  author={Ward-Cherrier, Benjamin and Pestell, Nicholas and Cramphorn, Luke and Winstone, Benjamin and Giannaccini, Maria Elena and Rossiter, Jonathan and Lepora, Nathan F},
  journal={Soft robotics},
  volume={5},
  number={2},
  pages={216--227},
  year={2018},
  publisher={Mary Ann Liebert, Inc. 140 Huguenot Street, 3rd Floor New Rochelle, NY 10801 USA}
}

@inproceedings{hogan2021seeing,
  title={Seeing Through your Skin: Recognizing Objects with a Novel Visuotactile Sensor},
  author={Hogan, Francois R and Jenkin, Michael and Rezaei-Shoshtari, Sahand and Girdhar, Yogesh and Meger, David and Dudek, Gregory},
  booktitle={Proceedings of the IEEE/CVF Winter Conference on Applications of Computer Vision},
  pages={1218--1227},
  year={2021}
}

@inproceedings{hellebrekers2020localization,
  title={Localization and Force-Feedback with Soft Magnetic Stickers for Precise Robot Manipulation},
  author={Hellebrekers, Tess and Zhang, Kevin and Veloso, Manuela and Kroemer, Oliver and Majidi, Carmel},
  booktitle={2020 IEEE/RSJ International Conference on Intelligent Robots and Systems (IROS)},
  pages={8867--8874},
  year={2020},
  organization={IEEE}
}

@inproceedings{yin2020closing,
  title={Closing the Loop with Liquid-Metal Sensing Skin for Autonomous Soft Robot Gripping},
  author={Yin, Jessica and Hellebrekers, Tess and Majidi, Carmel},
  booktitle={2020 3rd IEEE International Conference on Soft Robotics (RoboSoft)},
  pages={661--667},
  year={2020},
  organization={IEEE}
}

@article{navarro2021proximity,
  title={Proximity Perception in Human-centered Robotics: A Survey on Sensing Systems and Applications},
  author={Navarro, Stefan Escaida and M{\"u}hlbacher-Karrer, Stephan and Alagi, Hosam and Zangl, Hubert and Koyama, Keisuke and Hein, Bj{\"o}rn and Duriez, Christian and Smith, Joshua},
  journal={IEEE Transactions on Robotics},
  year={2021}
}

@article{zhao2016optoelectronically,
  title={Optoelectronically innervated soft prosthetic hand via stretchable optical waveguides},
  author={Zhao, Huichan and O’Brien, Kevin and Li, Shuo and Shepherd, Robert F}
}

@misc{intelrealsense, title={Optimizing the Intel RealSense LiDAR Camera L515 Range}, url={https://www.intelrealsense.com/optimizing-the-lidar-camera-l515-range/}, journal={Intel® RealSense™ Depth and Tracking Cameras}, publisher={Intel}, year={2020}, month={Oct}}

@article{bradski2000opencv,
  title={OpenCV},
  author={Bradski, Gary and Kaehler, Adrian},
  journal={Dr. Dobb’s journal of software tools},
  volume={3},
  year={2000}
}

@inproceedings{donlon2018gelslim,
  title={Gelslim: A high-resolution, compact, robust, and calibrated tactile-sensing finger},
  author={Donlon, Elliott and Dong, Siyuan and Liu, Melody and Li, Jianhua and Adelson, Edward and Rodriguez, Alberto},
  booktitle={2018 IEEE/RSJ International Conference on Intelligent Robots and Systems (IROS)},
  pages={1927--1934},
  year={2018},
  organization={IEEE}
}

@article{yuan2017gelsight,
  title={Gelsight: High-resolution robot tactile sensors for estimating geometry and force},
  author={Yuan, Wenzhen and Dong, Siyuan and Adelson, Edward H},
  journal={Sensors},
  volume={17},
  number={12},
  pages={2762},
  year={2017},
  publisher={Multidisciplinary Digital Publishing Institute}
}

@article{lambeta2020digit,
  title={Digit: A novel design for a low-cost compact high-resolution tactile sensor with application to in-hand manipulation},
  author={Lambeta, Mike and Chou, Po-Wei and Tian, Stephen and Yang, Brian and Maloon, Benjamin and Most, Victoria Rose and Stroud, Dave and Santos, Raymond and Byagowi, Ahmad and Kammerer, Gregg and others},
  journal={IEEE Robotics and Automation Letters},
  volume={5},
  number={3},
  pages={3838--3845},
  year={2020},
  publisher={IEEE}
}

@inproceedings{padmanabha2020omnitact,
  title={Omnitact: A multi-directional high-resolution touch sensor},
  author={Padmanabha, Akhil and Ebert, Frederik and Tian, Stephen and Calandra, Roberto and Finn, Chelsea and Levine, Sergey},
  booktitle={2020 IEEE International Conference on Robotics and Automation (ICRA)},
  pages={618--624},
  year={2020},
  organization={IEEE}
}

@inproceedings{rocha2017soft,
  title={Soft-matter sensor for proximity, tactile and pressure detection},
  author={Rocha, Rui and Lopes, Pedro and de Almeida, Anibal T and Tavakoli, Mahmoud and Majidi, Carmel},
  booktitle={2017 IEEE/RSJ International Conference on Intelligent Robots and Systems (IROS)},
  pages={3734--3738},
  year={2017},
  organization={IEEE}
}

@inproceedings{hellebrekers2018liquid,
  title={Liquid metal-microelectronics integration for a sensorized soft robot skin},
  author={Hellebrekers, Tess and Ozutemiz, Kadri Bugra and Yin, Jessica and Majidi, Carmel},
  booktitle={2018 IEEE/RSJ International Conference on Intelligent Robots and Systems (IROS)},
  pages={5924--5929},
  year={2018},
  organization={IEEE}
}

@incollection{sun_improving_2018,
	address = {Cham},
	title = {Improving {Grasp} {Performance} {Using} {In}-{Hand} {Proximity} and {Contact} {Sensing}},
	volume = {816},
	booktitle = {Robotic {Grasping} and {Manipulation}},
	publisher = {Springer International Publishing},
	author = {Patel, Radhen and Curtis, Rebeca and Romero, Branden and Correll, Nikolaus},
	editor = {Sun, Yu and Falco, Joe},
	year = {2018},
	doi = {10.1007/978-3-319-94568-2_9},
	note = {Series Title: Communications in Computer and Information Science},
	pages = {146--160},

}

@book{konstantinova_force_2015,
	title = {Force and {Proximity} {Fingertip} {Sensor} to {Enhance} {Grasping} {Perception}},
	author = {Konstantinova, Jelizaveta and Stilli, Agostino and Althoefer, Kaspar},
	month = sep,
	year = {2015},
	doi = {10.1109/IROS.2015.7353659}
}

@article{ruggiero2018nonprehensile,
  title={Nonprehensile dynamic manipulation: A survey},
  author={Ruggiero, Fabio and Lippiello, Vincenzo and Siciliano, Bruno},
  journal={IEEE Robotics and Automation Letters},
  volume={3},
  number={3},
  pages={1711--1718},
  year={2018},
  publisher={IEEE}
}

@article{shimonomura2019tactile,
  title={Tactile image sensors employing camera: A review},
  author={Shimonomura, Kazuhiro},
  journal={Sensors},
  volume={19},
  number={18},
  pages={3933},
  year={2019},
  publisher={Multidisciplinary Digital Publishing Institute}
}

\end{document}